\documentclass[10pt,twocolumn,letterpaper]{article}

\usepackage{wacv}
\usepackage{times}
\usepackage{epsfig}
\usepackage{graphicx}
\usepackage{tabularx}
\usepackage{amsmath}
\usepackage{amssymb}
\usepackage{booktabs}
\usepackage{mathtools}
\usepackage{multirow}
\usepackage{pifont}
\usepackage{adjustbox}
\usepackage{bm}
\usepackage[square,sort,comma,numbers]{natbib}
\newcommand{\cmark}{\ding{51}}
\newcommand{\xmark}{\ding{55}}
\usepackage[font=small,labelfont=bf]{caption}

\wacvfinalcopy

\ifwacvfinal
\usepackage[breaklinks=true,bookmarks=false]{hyperref}
\else
\usepackage[pagebackref=true,breaklinks=true,colorlinks,bookmarks=false]{hyperref}
\fi

\DeclareMathOperator{\Sim}{sim}

\begin{document}

\title{Unsupervised Video Representation Learning by Bidirectional Feature Prediction}

\author{Nadine Behrmann\\
Bosch Center for Artificial Intelligence\\
{\tt\small nadine.behrmann@de.bosch.com}
\and
Juergen Gall\\
University of Bonn\\
{\tt\small gall@iai.uni-bonn.de}
\and
Mehdi Noroozi\\
Bosch Center for Artificial Intelligence\\
{\tt\small mehdi.noroozi@de.bosch.com}
}

\maketitle

\begin{abstract}
  This paper introduces a novel method for self-supervised video representation learning via feature prediction. In contrast to the previous methods that focus on future feature prediction, we argue that a supervisory signal arising from unobserved past frames is complementary to one that originates from the future frames. The rationale behind our method is to encourage the network to explore the temporal structure of videos by distinguishing between future and past given present observations.  We train our model in  a contrastive learning framework, where joint encoding of future and past provides us with a comprehensive set of temporal hard negatives via swapping. We empirically show that utilizing both signals enriches the learned representations for the downstream task of action recognition. It outperforms independent prediction of future and past.
\end{abstract}

\section{Introduction}
\begin{figure*}
\centering
\includegraphics[width=\textwidth]{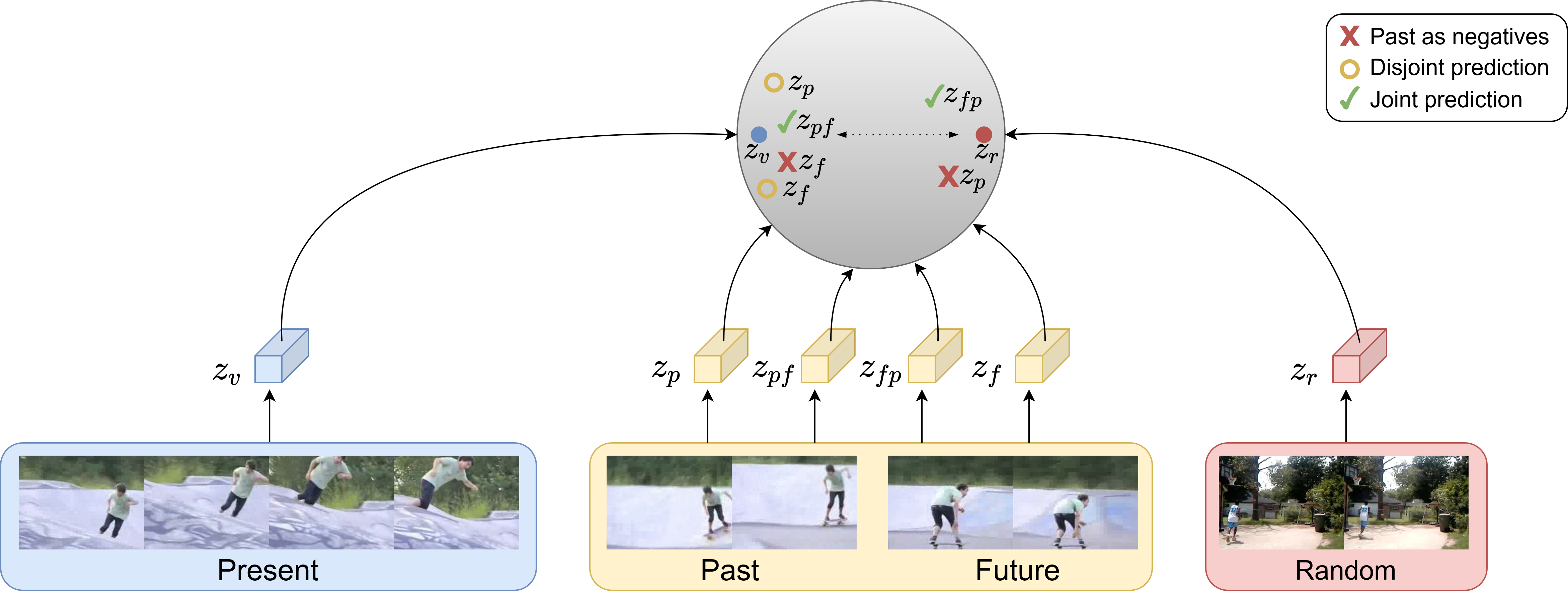}
\caption{
{\bf Effect of Past and Future Prediction on the Feature Space.}
The proximity of the features depicted in the gray feature space indicates their (dis)similarity. An ideal embedding space has two properties. I) Present, past, and future are similar II) Past and future are distinguishable given present. 
We consider three approaches:
The {\bf disjoint prediction} of past and future (yellow circles) only satisfies (I).
Predicting past $z_p$ rather than future $z_f$ requires similar attributes and consequently does not enhance the representations.
A naive extension is to use {\bf past as negatives} (red crosses), which only satisfies (II).
However, this leads to a degenerate solution where past is dissimilar from present and future.
In the former we want the features of both future and past to be distinguishable from random, while in the later we want to distinguish between future and past. 
{\bf Joint prediction} (green ticks) allows us to exploit both supervisory signals. The combination of past and future in the right order should be distinguishable from random as well as from the wrong order. In particular, $z_v$ and $z_{pf}$ should be close, promoting the similarity of past, present and future (I), while the wrong order of future and past $z_{fp}$ should be distinguishable (II), encouraging temporally structured representations. 
}
\label{fig:video_example}
\end{figure*}
Videos provide a rich source of information for visual understanding. They generously reveal to our machines how objects interact with each other and the environment in the real world. Nevertheless, the task of representing high-level abstractions from videos is essential to address a large and sophisticated set of downstream tasks tied to videos, \eg~temporal segmentation~\cite{Richard:2017:WSA}. %\cite{THUMOS14}. %  

Self-supervised learning (SSL) has recently established a promising direction for this purpose. A well known motivation for SSL from a practical point of view is to alleviate the cost and error of the labeling process. Moreover, learning generalizable and optimal representations can not be taken for granted in a supervised setting, especially when representing a complex source of information like videos. For instance, it has been shown that action labels are predictable from a single frame to an acceptable extent~\cite{Fouhey:2018:FLV}, providing a relatively weak source of supervision signal for representation learning as the network is not forced to explore temporal information of videos. In contrast to the image domain where the gap between self-supervised and supervised representation learning has been shrunk remarkably~\cite{Chen:2020:SimCLR, Asano:2020:SLC}, self-supervised video representation learning is still behind supervised learning even regarding relatively simple tasks such as action recognition.

In contrast to images, the temporal structure and multi-modal nature of videos provide even more opportunities to construct pretext tasks. While recent work on self-supervised multi-modal video representation learning has shown to be very effective \citep{Piergiovanni:2020:ELo}, we are interested in RGB-only self-supervised video representation learning.  Besides the scientific value of pure vision based models, a practical motivation involves applications where the audio signal is not accessible, e.g. autonomous driving.

We introduce a novel SSL task via feature prediction. Previous approaches of learning from prediction have been limited to future observations.
They aim to train a model that takes a segment of a video clip as input and predicts some form of the contents of future frames~\cite{LeCun:2016:DMV}. Our main idea involves incorporating unobserved past and future jointly.  
Given a segment of a video as a present sequence, the question of what will happen in the future frames is comparable to asking what has happened in the past in terms of abstract factors of variation which the network needs to encode.

Nonetheless, utilizing both signals is not trivial. 
The feature prediction task involves maximizing mutual information approximated mostly via a contrastive loss that compares joint and product of the marginals of present and future/past distributions. The quality of learned representations is highly dependent on the set of negatives~\cite{Tschannen:2020:OMI}.
While random sequences form the basis for negatives, they are easily distinguishable from the matching pair via shortcuts such as low level statistics, edges, etc. 
To prevent these shortcuts, the key for a comprehensively challenging pretext task is to construct additional negatives that are hard to distinguish, encouraging the network to explore the structure of the data more intensively.
Disjoint prediction of past and future does not introduce a new set of negatives compared to individual prediction of future or past, see Figure~\ref{fig:video_example}.\footnote{Note that past prediction fits to our terminology as we use the term of prediction to associate an observed partition of data to the unobserved ones.} 

Instead, we propose to predict future and past jointly.
Our method establishes a connection between present frames with a pair of future and past frames. This allows us to incorporate the wrong order of future and past as hard negatives. Given present frames, the network should  distinguish the corresponding pair of future and past from not only randomly taken pairs but also the swapped future and past of the same video. Our experiments show that joint prediction of future and past outperforms disjoint prediction on several transfer learning benchmarks.

Our contributions include: 
1) We propose a novel bidirectional feature prediction pretext task for video representation learning. 
2) We extensively evaluate our proposed method on several transfer learning benchmarks showing its superiority to its future prediction counterpart.

\section{Related Work}
An important category of self-supervised learning methods involves dividing data into two segments, and training a convolutional neural network (CNN) that \textit{predicts} one part given the other. 
The key ingredients for these approaches are the design of a partitioning paradigm and the definition of a loss function via quantifying prediction.
For instance, the task of image colorization~\citep{Zhang:2016:colorful,Larsson:2016:color} proposes to divide the input image into $Lab$ channels and train a model that predicts $ab$ channels given $L$ channel. 
In the video domain, partitioning in temporal direction has been intensively explored ~\citep{Lotter:2017:PCN, LeCun:2016:DMV, Srivastava:2015:ULV, Vondrick:2016:GVS}.

\paragraph{Future prediction.}
A straight forward approach to quantify prediction is based on raw data, where the target distribution is fixed as true data distribution. 
A naive approach includes reconstruction losses, which are based on strong assumptions  and consequently do not achieve decent results.\footnote{For instance, $\ell_2$ assumes the data follows a Gaussian distribution while cross-entropy provides a discrete approximation of distribution.}
To eliminate these limitations, an adversarial loss has been employed in~\citep{Pathak:2016:context, LeCun:2016:DMV}. 
However, the supervisory signals arising from true data distributions suffer from ambiguity and require the network to devote substantial capacity to model a data distribution, which is not necessarily the optimal solution for representation learning. 
Addressing this problem, \citet{Vondrick:2016:AVR} suggest to predict the representation of future frames instead of pixel values, and use a mixture model to handle the multi-modal distribution of future representations.
Recently, predicting features of the future video blocks in a contrastive fashion has gotten great attention~\citep{Han:2019:DPC, Han:2020:MemDPC}, achieving remarkably better performance. 
The idea is to perform prediction in a learnable transformation of raw data trained jointly with the prediction quantification.
More specifically,~\citet{Han:2019:DPC} train a 3D CNN that takes a sequence of video blocks and predicts the features of a future block. They apply the same backbone to extract features from present and future blocks and quantify prediction via the \textit{InfoNCE} loss~\citep{Oord:2018:CPC}, which involves maximizing an estimate of mutual information \citep{Poole:2019:OVB} and has been successfully applied in multiple domains~\citep{Chen:2020:SimCLR}.
In an extension of their work, \citet{Han:2020:MemDPC} very recently propose a memory-augmented version and consider both optical flow and RGB videos as input modalities.
In contrast to these methods based on \textit{unidirectional} feature prediction, we propose \textit{bidirectionally} predicting future and past features and adopt the InfoNCE loss for prediction quantification.

\paragraph{InfoNCE loss.}
\citet{Oord:2018:CPC} propose contrastive predictive coding applied to images by sequentially predicting representations of image patches.
\citet{Bachman:2019:LRM} and \citet{Tian:2019:CMC} maximize the mutual information of different views of the same image via InfoNCE.
\citet{Chen:2020:SimCLR} improve the InfoNCE loss in several ways. First, they show that applying a learnable nonlinear transformation before the contrastive loss is superior to a linear transformation. Furthermore, they demonstrate that normalizing the representations, using a temperature parameter and a large batch size,~\ie~the number of negatives, are beneficial. 
\citet{He:2019:MoCo} decouple the number of negatives from the batch size by introducing a method to effectively employ a memory bank for storing the negatives.

\paragraph{Pretext tasks for video representation learning.}
Different approaches for video representation learning orthogonal to future prediction have been explored.
While image-level pretext tasks can be extended to videos, \eg ~\citep{Jing:2018:SSS, Kim:2019:STC}, the temporal dimension of videos provides the opportunity of a more comprehensive set of pretext tasks.
These include verification of the temporal order of frames~\citep{Misra:2016:SAL, Lee:2017:URL, Fernando:2017:SSV, Wei:2018:LUA}, predicting the order of clips instead of frames~\citep{Xu:2019:VCOP}, egomotion~\citep{Jayaraman:2015:LIR, Agrawal:2015:LSM}, temporal coherency in video~\citep{Wang:2015:ULV, Isola:2016:LVG, Jayaraman:2016:SSF, Lai:2019:CVF} and predicting video speed~\citep{Benaim:2020:SpeedNet, Epstein:2020:Oops}. \citet{Vondrick:2018:TEC} show that the temporal coherency of color enables the learning of a tracking algorithm whereas \citet{Wang:2019:LCC} use the cycle-consistency of time to learn correspondence.

\section{Method}
\label{sec:method}
\begin{figure*}[t]
	\centering
    \includegraphics[width=\textwidth]{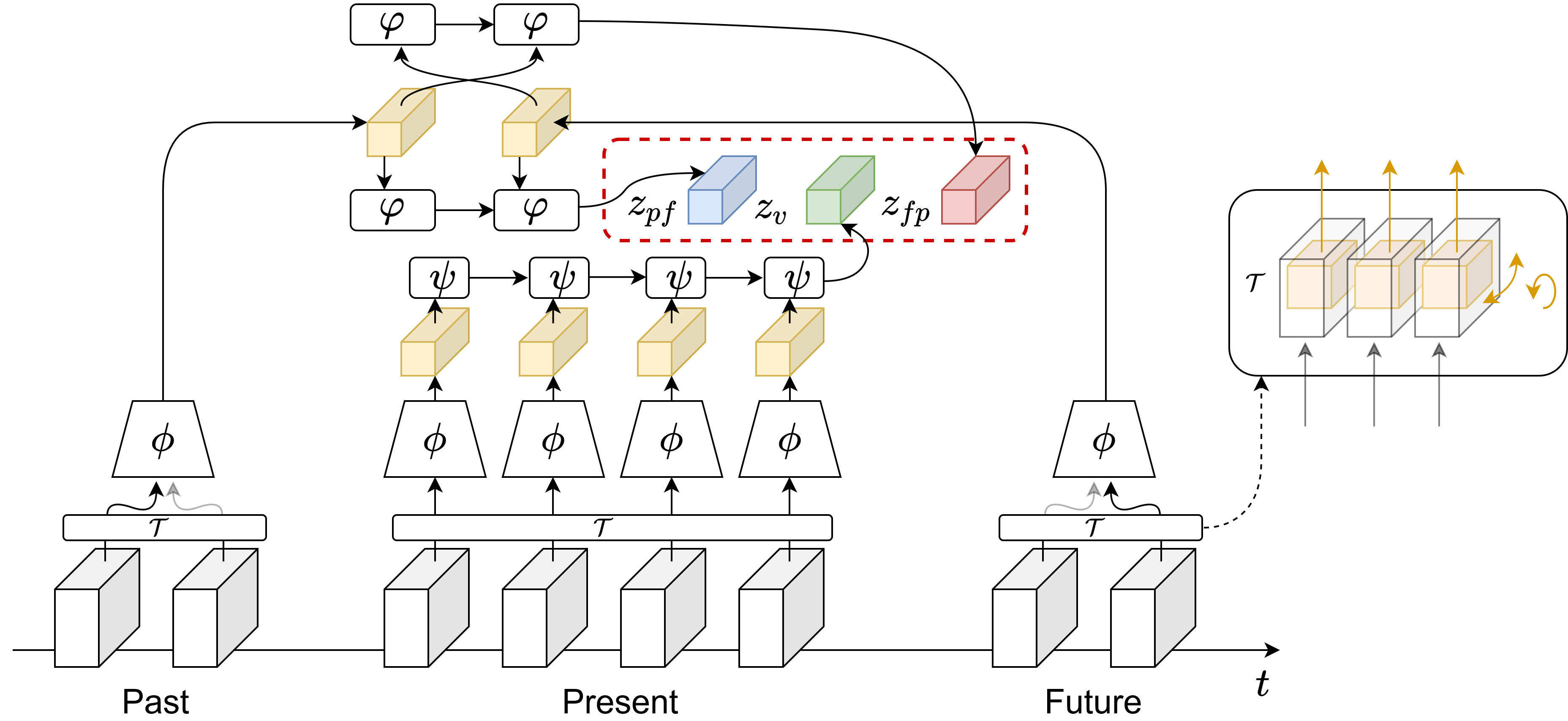}
	\caption[Bidirectional Feature Prediction]{
	{\bf Bidirectional Feature Prediction.}
	First we apply spatial transformations $\mathcal{T}$ to past, present and future independently, these can include random crop, horizontal flip and rotation.
	Then a 3D CNN $\phi$ is applied to the non-overlapping blocks of frames. 
	To obtain a video representation $z_v$ an aggregation function $\psi$ is applied to the sequence of features.
	We randomly sample a past and a future block, which is indicated by the black and gray arrows, and apply an aggregation function $\varphi$ to acquire a merged representation $z_{pf}$ of past and future.
	A temporal hard negative $z_{fp}$ is obtained by applying $\varphi$ to the swapped features of future and past.
	To maximize the agreement of $z_v$ and $z_{pf}$ we employ the InfoNCE loss where we incorporate $z_{fp}$ as temporal hard negative in the denominator.
	}
	\label{fig:bidirectional_feature_prediction}
\end{figure*}

Our method involves predicting unseen past and future features of a video sequence.
We employ a variant of Noise Contrastive Estimation, \textit{InfoNCE}, in which the network is asked to distinguish the positive pair from a set of negatives.
More precisely, given a video clip we divide the video frames into three partitions, $X~=~(P, V, F)$, each element refers to past, present, and future respectively. We then construct the positive and negative pairs to train InfoNCE by exploiting the joint representations of $(P,F)$.

\paragraph{Past and future feature prediction.}
First, we explore how the two complementary supervision signals can be incorporated into a comprehensively challenging pretext task.
This seemingly easy objective proves to be non-trivial and deserves an in-depth discussion.
A naive approach to combine past and future prediction will be in a disjoint fashion.
That is, to view the task of past and future prediction independently in a unidirectional manner by simply adding the respective losses. This is equivalent to encouraging distinguishable features of both future and past from random given present.
Unfortunately, this straightforward approach does not achieve decent results, see Table~\ref{table:comparison_state_of_the_art}.
We conjecture that representations which are able to predict the future are sufficient to predict the past and thus the added losses do not stimulate the network to enhance the representations.

Instead, to encourage the model to explore the temporal structure further, we construct a pretext task that involves distinguishing between future and past, given present.
We did not achieve reasonable results by explicitly classifying future, past, and random pairs. Alternatively, we use the InfoNCE loss and implicitly exploit this signal.\footnote{One reason could be that the network cooperates with the classifier to embed some shortcuts in the feature space. The InfoNCE does not include the classifier and puts more load on the feature extraction network, and yields higher performance.}
An apparent choice is to include past features as negative pairs, while future features form the positive pairs.
However, this is not an appropriate approach. 
Essentially, past features should encode similar high level abstractions as those of the future,~\eg~underlying action. Having past as negatives entails dissimilarity between representations of the past and present/future, see Figure~\ref{fig:video_example}, resulting in a degenerate solution that removes such high level abstractions shared across past, present, and future from the representations. Our experiment termed \textit{Past as negatives} in Table~\ref{table:comparison_state_of_the_art} confirms the poor quality of the resulting representations of this approach. A desired solution favors distinguishable features of future and past given present. To this end, we propose joint prediction of future and past. Our model takes a pair of past/future as input, which allows temporal hard negatives of the wrong order of future/past.
Distinguishing between past/future and future/past requires the network to encode temporal structure shared across the video 
such that matching temporal orders can be detected. Our approach encourages the features of future and past to be  distinguishable from random as well as each other given present.

\paragraph{Network architecture.}
For fair comparisons, we use a 2D3D version of ResNet18 as in~\citep{Han:2019:DPC} that consists of 3D convolutions only in the last two layers.
The video sequence is divided into non-overlapping blocks of frames. We extract convolutional features from the blocks via a shared 2D3D-ResNet18, $\phi$.
To create the final features, we aggregate the extracted convolutional features via one-layer Convolutional Gated Recurrent Units (ConvGRU) with kernel size $1$. We found that using a separate aggregation function for past/future blocks achieves slightly better results, see supplementary material.
Consequently, we construct the final feature of present blocks as $z_v=\left(\psi \circ \phi\right)(V)$, and past/future blocks as $z_{pf}=\left(\varphi \circ \phi\right)(P,F)$, where $\psi$, $\varphi$ denote aggregation functions of present and past/future blocks respectively. 
As $\varphi$ is non-symmetric, we can apply it to the reverse order of future/past to obtain a temporal hard negative $z_{fp}=\left(\varphi \circ \phi\right)(F,P)$, which will be discussed below.
Figure~\ref{fig:bidirectional_feature_prediction} shows all components of our method.

\subsection{Contrastive Loss} 
For a given video, we build a set $\mathcal{P}$ consisting of positive future and past pairs. 
For each positive pair $(P,F)\in\mathcal{P}$ we consider a set $\mathcal{D}_{(P,F)}$ containing the positive pair itself and all its negatives, including $z_{fp}$. As suggested in \citep{Chen:2020:SimCLR}, we use a small MLP head $f$ to map the representations to the space in which the contrastive loss is applied, denoted by $f_v=f(z_v)$ and $f_{pf}=f(z_{pf})$, respectively.
We use the cosine similarity $\Sim(u,v)=\frac{u^Tv}{\|u\|\|v\|}\cdot \frac{1}{\tau}$ with a temperature parameter $\tau$ to compare present features with those obtained from future and past,

\begin{equation}
\label{eq:InfoNCEloss}
    \mathcal{L} =\sum_{(P,F) \in \mathcal{P} } {}-\log\left(\frac{\exp(\Sim(f_v,f_{pf}))}{\sum_{(\hat P, \hat F) \in \mathcal{D}_{(P,F)}} \exp(\Sim(f_v,f_{pf}))} \right).
\end{equation}
We discuss the process of constructing positive and negative pairs in the following.

\textbf{Positives.} $\mathcal{P}$ denotes a set of positive past/future blocks. We obtain multiple positive pairs per video by selecting a random block from past and future blocks respectively. If there are $m$ past and future blocks per video, we build $m^2$ positive pairs per video, see Figure~\ref{fig:bidirectional_feature_prediction}.

\textbf{Easy negatives.} We obtain easy negatives by sampling all possible combinations of past/future and future/past blocks from other videos in a batch. If there are $m$ past and future blocks per video, a batch size of $n$ provides us with $2m^2(n-1)$ easy negatives.

\textbf{Temporal hard negatives.} Temporal hard negatives are obtained via swapping the order of past and future.
Consequently, we obtain the same number of temporal hard negatives per video as the number of positives.
Each set $\mathcal{D}_{(P,F)}$ contains all temporal negatives.

\textbf{Independent augmentation vs. spatial negatives.}
\citet{Han:2019:DPC} use spatial negatives in their method where they decompose the $4\times 4 \times 256$ convolutional feature map along the depth direction into $16$  feature vectors, and treat all of those that do not match the corresponding spatial location as negatives. 
We propose to simply apply independent augmentations to $P,V,F$, obtaining a single $256$ dimensional feature vector via global average pooling. Our experiments show that simple independent augmentations achieve higher performance than complex spatial negatives. This is inline with recent observations made by~\cite{Misra:2019:PIRL}. Spatial negatives aim to learn features variant to the spatial location. That is while independent augmentations aims to learn features invariant to the spatial location. Moreover, the major drawback of spatial negatives is the following: They encourage the feature vectors to represent local descriptors. The feature vectors should be distinguishable across spatial locations of the feature map since they are injected as negatives in the loss function. This assumption might be useful in the early layers where the receptive fields of neurons are small. However, in the later layers where the receptive field grows, a global feature is favorable. Our experiments in Table~\ref{table:eval_kinetics} shows that indeed the features learned via spatial negatives transfer poorly to the downstream task in the later layers.

\subsection{Connection to Mutual Information Maximization}
The InfoNCE loss has been shown to be a lower bound of mutual information \citep{Poole:2019:OVB}.
From this point of view, our loss in Eq.~\ref{eq:InfoNCEloss} can be interpreted as mutual information maximization of the features extracted by $\phi$:
\begin{align}
\max_\phi \quad I(\phi(V), \phi(P,F)).
\label{eq:1}
\end{align}

A correct lower bound is obtained when the negative pairs are built via sampling from the product of the marginals. Our temporal hard negatives are not sampled independently from the present blocks, imposing an incorrect approximation of the mutual information. However, this is not counterproductive. As it has been recently shown~\citep{Tschannen:2020:OMI}, the effectiveness of the InfoNCE loss for learning representation is not necessarily tied to the accuracy of mutual information estimation. Structured hard negatives which are not sampled from the product of the marginals contribute to the quality of the learned representations more than an accurate estimation of mutual information. We could not achieve decent results without temporal hard negatives, see supplementary material, confirming the same argument.

\section{Experiments}
\begin{table*}[t]
    \centering
    \begin{tabular}{lllcl|cc}
    \toprule
    \multicolumn{5}{c|}{Self-Supervised Methods}&\multicolumn{2}{c}{top1 Accuracy}\\
    Method & Architecture & Pretraining Dataset & Spatial Negatives & Ref & UCF101 & HMDB51\\
    \midrule
    Random Initialization & 2D3D-Resnet18 & - &  & - & $54.4$ & $31.9$\\
    Random Initialization & R3D & - &  & \citep{Xu:2019:VCOP} & $54.4$ & $21.5$ \\
    Random Initialization & S3D-G & - &  & \citep{Benaim:2020:SpeedNet} & $73.8$ & $46.4$ \\
    \midrule
    Shuffle\&Learn \citep{Misra:2016:SAL} & CaffeNet & UCF101/HMDB51 &  & \citep{Han:2019:DPC} & $50.2$ & $18.1$ \\
    OPN \citep{Lee:2017:URL} & VGG-M-2048 & UCF101/HMDB51 &  & \citep{Han:2019:DPC} & $59.8$ & $23.8$ \\
    VCOP \citep{Xu:2019:VCOP} & R3D & UCF101 &  & \citep{Xu:2019:VCOP} & $64.9$ & $29.5$\\
    3DRot \citep{Jing:2018:SSS} & 3D-Resnet18 & Kinetics-600 &  & \citep{Han:2019:DPC} & $62.9$ & $33.7$\\
    3D-ST-Puzzle \citep{Kim:2019:STC} & 3D-Resnet18 & Kinetics-400 &  & \citep{Han:2019:DPC} & $65.8$ & $33.7$\\
    DPC \citep{Han:2019:DPC} & 2D3D-Resnet34 & Kinetics-400 &  & \citep{Han:2019:DPC} & $75.7$ & $35.7$ \\
    SpeedNet \citep{Benaim:2020:SpeedNet} & S3D-G & Kinetics-400 &  & \citep{Benaim:2020:SpeedNet} & $81.1$ & $48.8$ \\
    \midrule
    Future prediction & 2D3D-Resnet18 & UCF101 & \cmark & - & $61.3$ & - \\
    Past prediction & 2D3D-Resnet18 & UCF101 & \cmark & - & $60.1$ & - \\
    Disjoint prediction & 2D3D-Resnet18 & UCF101 & \cmark & - & $60.1$ & - \\
    Past as negatives & 2D3D-Resnet18 & UCF101 & \cmark & - & $57.6$ & - \\
    Ours & 2D3D-Resnet18 & UCF101 & \cmark & - & $\bm{63.6}$ & - \\
    \midrule
    Future prediction & 2D3D-Resnet18 & Kinetics-400 & \cmark & - & $65.9$ & $35.3$ \\
    Ours & 2D3D-Resnet18 & Kinetics-400 & \xmark & - & $\bm{66.4}$ & $\bm{45.3}$\\
    \bottomrule
    \end{tabular}
    \caption{
    {\bf Finetuning on UCF101 and HMDB51.}
    The second block shows methods with different architectures. The  third and fourth blocks are pretrained on UCF101 and Kinetics-400 respectively. For comparison we report results with random initialization in the first block. 
    The Ref column indicates the source the values were taken from. Values without a reference were generated by us.
    Future prediction refers to the DPC method trained with our implementation details, while Past prediction denotes the DPC method applied to predicting past features. Disjoint prediction constitutes the added losses of past and future prediction and Past as negatives refers to InfoNCE trained with future features as positives and past features as negatives.
    While we observe moderate improvements of our method on UCF101, it significantly boots the performance on HMDB51.
    }
    \label{table:comparison_state_of_the_art}
\end{table*}{}
Following the common practice in self-supervised video representation learning, we evaluate our approach on the downstream task of action recognition.
During the self-supervised pretraining stage we discard the labels of the dataset and train a model using Eq.~\ref{eq:InfoNCEloss}.
Note that although we drop the class labels, a supervision bias remains in the dataset, \eg the videos in Kinetics-400 are temporally trimmed and carefully selected. However, this is a widely adopted approach and we follow previous works in order to do fair comparisons.
To evaluate the learned representation, we transfer the pretrained network, including the 2D3D-Resnet18 backbone $\phi$ and aggregation function $\psi$. 
We add a linear layer on top of the resulting feature vector and evaluate our network for action recognition on labeled data.
We follow two paradigms: 1) We \textit{finetune} the entire network with randomly initialized linear classification layer for the task of action recognition, using a reduced learning rate for pretrained layers.
2) Additionally, we evaluate the quality of the representations layer-wise by \textit{freezing} the pretrained network up to some layer, and training the remaining from scratch. This evaluation was primarily proposed by~\cite{jigsaw} in the image domain. 
Our experiment on Kinetics-400 allows a more comprehensive evaluation of the quality of learned representations as it gives a better insight into the learned representations.
It is well known that the early layers of CNNs extract general and local features whereas later layers are more specific to the training task and dataset~\cite{Yosinski:2014:HTF}.
A better performance on the downstream task of the later layers indicates higher correlation of pretext and downstream task.

\paragraph{Implementation details.}
We use similar self-supervised learning and evaluation settings to \citep{Han:2019:DPC}, our most important baseline. We extract $8$ blocks of $5$ frames from a video sequence and split them in the following way. The $4$ middle blocks are used as present video sequence, while we sample single past and future blocks at different temporal distances to the present from the remaining blocks.
In this case, we have $4$ positive pairs per video and $4$ corresponding negative pairs.
During self-supervised learning we apply random crop and horizontal flip, as well as frame-wise color jittering and random downsampling of the frame rate, 
where we sample video frames with a random stride of at most $3$.
For finetuning we keep the random crop and horizontal flip, but apply consistent color jittering at the video-level and sample frames with a constant stride of $3$.
In the case of spatial negatives, the spatial transformations crop and horizontal flip are applied consistently to the entire video.  As discussed above, we propose to augment past, present, and future video sequences independently, resulting in more robust representations.
We conduct experiments with both settings.
Note that spatial negatives can not be combined with independent spatial transformations since they require correspondence. 
During self-supervised training, we map the features to the contrastive space using a small MLP head, consisting of $256$ hidden units and ReLU activation function. After self-supervised training, the MLP is discarded and only the backbone and aggregation function are transferred to the downstream task. For the downstream task, we use the same input structure of video blocks as during SSL training.
For both self-supervised training and finetuning, we use a batchsize of $64$ and the Adam optimizer \citep{Kingma:2015:Adam} with a learning rate of $10^{-3}$ and weight decay $10^{-5}$. We reduce the learning rate by a factor of $10$ when the validation loss plateaus.
For the layer-wise evaluations we use SGD with an initial learning rate of $0.1$, a weight decay of $10^{-3}$, a momentum of $0.9$ and a learning rate scheduler.
For SSL training, we train the network for 100 and 470 epochs on Kinetics-400 and UCF101, respectively. For finetuning, we follow \citep{Han:2019:DPC} and fix the number of epochs to 300 and reduce the learning rate to $10^{-4}$ for the pretrained weights.
During inference, we divide videos from the validation set into blocks of $5$ frames and construct half-length overlapping sequences of blocks that are fed to the model. We remove augmentations and only use center crop. We average the softmax probabilities to obtain the final classification scores.

\subsection{Finetuning on UCF and HMDB}
A most widely used framework of self-supervised learning involves obtaining an initialization from a large scale unlabeled dataset, and finetuning on a small annotated target dataset. 
We consider UCF101~\citep{Soomro:2012:UCF} and HMDB51~\citep{Kuehne:2011:HMDB} as the standard benchmarks in this domain. We report numbers on split 1 for both datasets.

\paragraph{Pretraining on UCF.} 
The main motivation of self-supervised learning is to take advantage of large scale unlabeled datasets. Therefore, pretraining on a small dataset such as UCF101 does not establish a plausible framework. Following the community, we aim to quickly validate the design choices of our bidirectional feature prediction task in the following experiments.
We show the results in the third block of Table~\ref{table:comparison_state_of_the_art}. 
First, we compare our method to past and future prediction. The future prediction corresponds to the DPC method using our implementation details, which improves slightly over the original implementation~\cite{Han:2019:DPC}. The past prediction is the same task but the unobserved past frames are used instead of future frames. 
While both future prediction and past prediction learn representations that prove to be useful for action recognition, simply adding the two losses (\textit{Disjoint prediction}) does not improve the quality of the representations.
Another naive approach of incorporating both supervisory signals is to use past features as negatives in the InfoNCE loss, while only the future features are treated as positives, which implicitly requires the network to distinguish between past and future. However, this leads to  weaker representations by removing high level abstractions shared across the blocks, shown as \textit{Past as negatives} in Table~\ref{table:comparison_state_of_the_art}.
In contrast, our method based on bidirectional feature prediction improves the performance on the downstream task of action recognition and consequently validates our design choices.
While we outperform unidirectional prediction,
our method is slightly inferior to VCOP \citep{Xu:2019:VCOP}.
However, note that VCOP employs the R3D architecture that includes 3D convolutions in all layers whereas our 2D3D-Resnet18 uses 3D convolutions only in the last two layers. Their network consumes more memory and computations compared to ours. 

\paragraph{Pretraining on Kinetics.}
Next, we investigate the benefit of large-scale datasets.
We pretrain our model on the Kinetics-400 dataset \citep{Kay:2017:Kinetics}, and evaluate the learned representations by finetuning on UCF101 and HMDB51. 
The last block of Table~\ref{table:comparison_state_of_the_art} summarizes the results.
We outperform future prediction on both UCF101 and HMDB51, a dataset known to be notoriously difficult for action recognition. We achieve $45.3\%$ top1 accuracy on HMDB51 surpassing many previous self-supervised RGB-only methods by a significant margin. In the second block of Table~\ref{table:comparison_state_of_the_art}, we compare to various different methods using different network architectures and pretraining datasets. Note that 2D3D-Resnet34 and S3D-G are significantly deeper and computationally more expensive  than 2D3D-Resnet18.

\begin{table*}[t]
    \centering
    \begin{tabular}{lccccccc}
    \toprule
     \multirow{2}{*}{Model} & spatial & \multicolumn{6}{c}{top1 Accuracy on Kinetics-400}\\
     \cmidrule{3-7}
      & negatives &\texttt{conv\_1} & \texttt{res\_1} & \texttt{res\_2} & \texttt{res\_3} & \texttt{res\_4} & \texttt{agg} \\
     \midrule
     random  & - & $31.1$ & $26.6$ & $16.9$ &  $8.4$ &  $2.9$ &  $1.8$ \\
     \midrule
     DPC \citep{Han:2019:DPC}    & \cmark & $38.1$ & $37.8$ & $27.0$ & $19.3$ & $4.6$ &  $3.6$ \\
     \midrule
     Future prediction  & \cmark & $41.2$ & $40.5$ & $32.8$ & $19.8$ & $4.3$ & $2.4$ \\
     Ours sobel+crop+rot & \xmark & $41.6$ & $40.5$ & $34.4$ & $26.2$ & $9.8$ & $7.7$ \\
     \bottomrule
    \end{tabular}
    \caption{
    {\bf Layer-wise evaluations on Kinetics-400.} 
    Pretrained models are frozen up to a convolutional/residual layer \texttt{conv}/\texttt{res} or completely \texttt{agg}, remaining layers are trained from scratch. The augmentations involve random crop applied independently to each partition. Random sobel filtering and rotation with multiplicands of $90^{\circ}$ are independently applied to past and future blocks only. Methods with independent augmentations instead of spatial negatives are superior in later layers.
    }
    \label{table:eval_kinetics}
\end{table*}{}

\subsection{Layer-wise evaluation on Kinetics}
\begin{figure}
    \begin{center}
        \includegraphics[width=\linewidth]{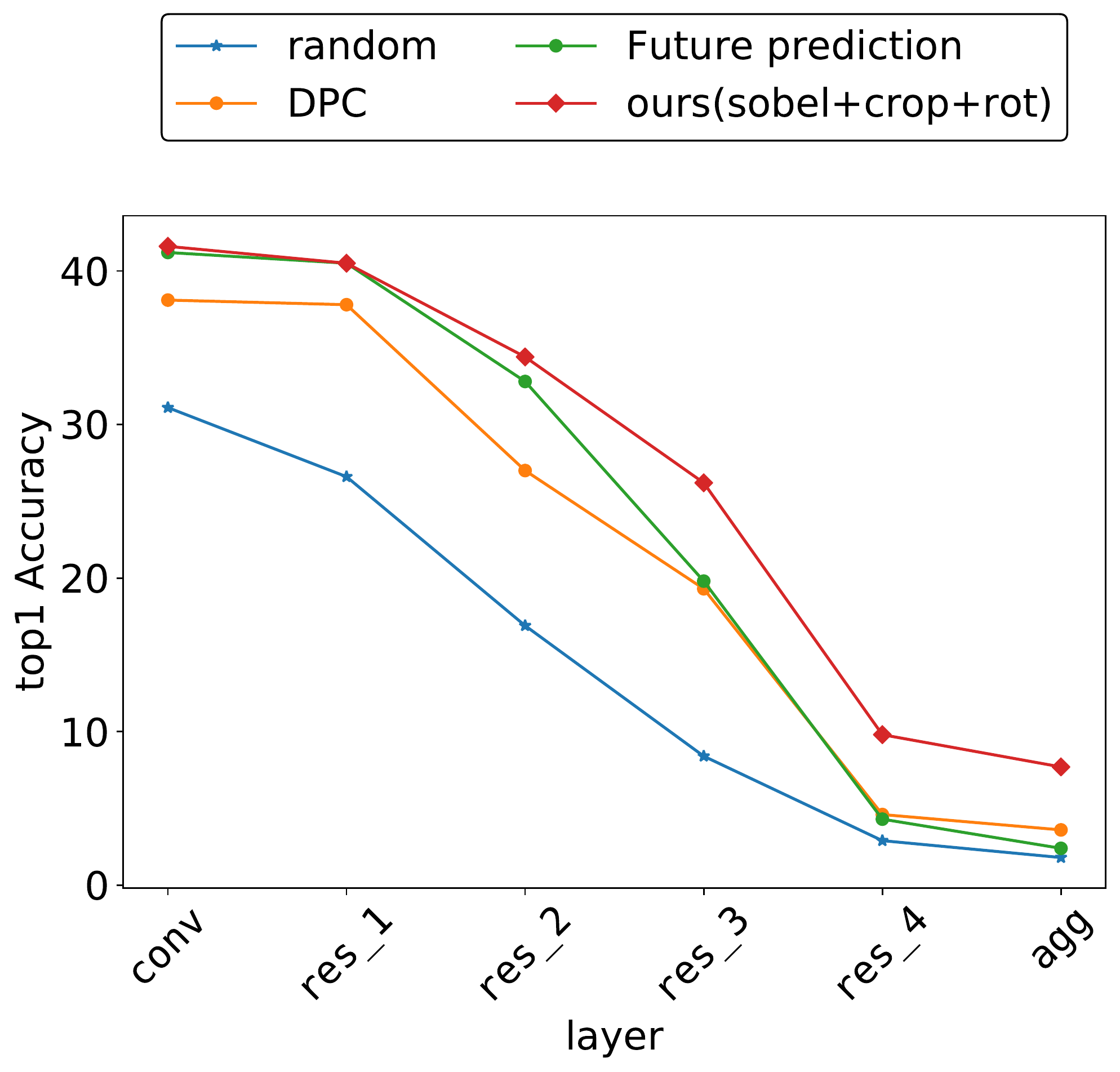}
    \end{center}
    \captionof{figure}{{\bf Layer-wise evaluations on Kinetics-400. }
    The corresponding results are shown in Table~\ref{table:eval_kinetics}.
    The pretrained networks are frozen up to some layer, the remaining are trained from scratch. 
    }
    \label{fig:frozen_layers}
\end{figure}

Finetuning is a well justified approach in practice. However, being prone to overfitting on the small target dataset, it does not necessarily establish a solid evaluation of the representations. To evaluate the learned representations more elaborately, we transfer the features of a pretrained model for action recognition on the Kinetics-400 dataset. We initialize multiple networks up to different layers with pretrained weights, and train only the remaining layers from scratch while the initialized layers are frozen. 
Provided enough training data, a better performance indicates higher quality representations of the initialized layers.
Table~\ref{table:eval_kinetics} and Figure~\ref{fig:frozen_layers} show the results when the initialized network is frozen up to the first convolutional (\texttt{conv}), $4$ residual (\texttt{res}) layers, and in the extreme case the aggregation function (\texttt{agg}).
Moreover, this evaluation allows us to investigate the effect of spatial negatives versus independent augmentations. The later layers with larger receptive fields represent global attributes of videos,~\eg~action class, whereas the earlier layers represent local attributes.
We obtain remarkably better performance in the later layers when the network is trained via independent transformations (our approach) compared with spatial negatives (Future prediction). The performance in early layers of both approaches is comparable. This observation confirms our argument in Section~\ref{sec:method}. 
Spatial negatives result in a representation that is variant with respect to the spatial region of the input and therefore encourage local feature descriptors, which may be useful in early layers where the receptive field is limited. Independent augmentations on the other hand lead to representations that are invariant to spatial transformations, allowing global feature descriptors that are more useful in later layers.

\begin{table}
\centering
    \begin{tabular}{lcc}
    \toprule
     \multirow{2}{*}{\# of ($P$, $V$, $F$)} & top1 Accuracy \\
      & on UCF101 \\
     \midrule
     $(2,4,2)$ & $63.6$ \\
     $(3,2,3)$ & $61.9$ \\
     $(2,2,2)$ & $\bm{63.8}$ \\
     $(1,2,1)$ & $62.9$ \\
     $(2,1,2)$ & $62.5$ \\
     \bottomrule
    \end{tabular}
    \vspace{0.5em}
    \captionof{table}{
    {\bf Ablation of the number of blocks.}
    Each triplet indicates the number of past, present, and future blocks per video in a batch. All present blocks are used to extract the present features. A pair of randomly taken single past and future blocks are used to construct past/future or future/past features.
    }
    \label{table:ablation_number_of_blocks}
\end{table}

\subsection{Ablation studies}
Table~\ref{table:ablation_number_of_blocks} shows the impact of different data partitioning paradigms into past, present, and future blocks. We pretrain different methods on split~1 of UCF101 and report the top1 accuracy obtained via finetuning on the same dataset. We follow the same set up as in the third block of Table~\ref{table:comparison_state_of_the_art}. We show each partition with a triplet referring to the number of blocks used to construct past, present, and future blocks respectively. We use all present blocks to construct present features, and randomly select a single block of past and future to build positive pairs of past/future and temporal negatives of future/past. The larger number of future and past blocks allows us to provide the InfoNCE loss with a larger set of positive and negative pairs. For instance, $(2,4,2)$ provides $4$ positive and negatives pairs per video in a batch. $(3,2,3)$ increases the difficulty of the task as more temporal hard negatives will be included in the loss function while the temporal receptive field of the present sequence is reduced. Reducing the number of present blocks while keeping the past and future blocks fixed does not change the quality of the representations significantly. However, further reducing the number of future and past blocks in $(1,2,1)$ or the number of present blocks in $(2,1,2)$ weakens the representations. 
The former reduces the number of temporal hard negatives which leads to a simpler task, while the later limits temporal information. This result indicates the effectiveness of temporal hard negatives as well as capability of our model to exploit temporal information.

\section{Conclusion}
In this paper we proposed a novel method for self-supervised video representation learning based on bidirectional feature prediction.
We showed how past and future prediction can be jointly incorporated in a contrastive learning framework and validated our design choices empirically.
We extensively evaluated our method via finetuning and layer-wise evaluations, outperforming unidirectional feature prediction  methods on the downstream task of action recognition.
\paragraph{Acknowledgment.}

JG has been supported by the Deutsche Forschungsgemeinschaft (DFG, German Research Foundation) under Germany’s Excellence Strategy - EXC 2070 –390732324, GA1927/4-1 (FOR 2535 Anticipating Human Behavior), and the ERC Starting Grant ARCA (677650).

{\small
\bibliographystyle{plainnat}
\bibliography{BidirectionalFeaturePrediction}
}

\cleardoublepage

\section*{Supplementary Material}

\subsection*{Ablation of our architecture}
As mentioned in Section~3 of the paper we found that using a different aggregation function (different instance of the ConvGRU) for past and future features while keeping the same 2D3D-Resnet18 backbone achieved better results. 
Here we want to investigate this further and compare the effect of using the same or different aggregation function or backbone for past and future features, \ie sharing weights of the feature extractor for the present representation $z_v$ and and past/future representation $z_{pf}$.
For this ablation we pretrain our models on UCF101 and evaluate the representations via finetuning. 
The results are shown in Table~\ref{table:ablation_architecture}.
When using the same aggregation function and the same or different backbone to extract past and future features, we observe a small performance drop compared to the best setting, whereas using both different backbones and different aggregation functions decreases the quality of the representation significantly.
We use the best performing setting for all of our experiments.
\begin{table}[h]
\centering
\begin{tabular}{ccc}
\toprule
 \multirow{2}{*}{backbone} & \multirow{2}{*}{agg} & top1 Accuracy \\ 
   & & on UCF101\\
 \midrule
 same & same & 62.4 \\
 different & same & 62.8 \\
 same & different & \textbf{63.6} \\
 different & different & 60.6 \\ 
\bottomrule
\end{tabular}
\vspace{0.5em}
 \caption{{\bf Ablation of our architecture.}
 }
 \label{table:ablation_architecture}
\end{table}
\subsection*{Temporal Negatives}
Next, we want to validate the effectiveness of our temporal negatives.
We train all methods on UCF101 using spatial transformations that are applied independently to the past, present and future blocks. For evaluation we finetune on UCF101.
All methods in Table~\ref{table:temporal_negatives} that do not employ temporal negatives fail, suggesting that the learned representations are bad initializations. 
Since neither temporal nor spatial negatives are used here, only random video sequences are considered as negatives and the InfoNCE loss in Eq.~1 of the paper gives a lower bound on the mutual information.
Adding temporal negatives on the other hand enables our method to learn a representation that is useful for action recognition.
These experiments confirm the observation in Section~3.2 that a structured set of hard negatives which are not sampled from the marginal distributions are more effective for representation learning than an accurate approximation of the mutual information.

\begin{table}
\centering
\begin{tabular}{lccc}
\toprule
 \multirow{2}{*}{Model} & temporal & spatial & top1 Accuracy \\ 
  & negatives & augmentations & on UCF101\\
\midrule
 Random Init & - & - & 54.4 \\ 
 \midrule
 Ours & \xmark & - & 48.2 \\
 Ours & \xmark & crop & 47.3 \\ 
 Ours & \xmark & crop + flip & 47.5 \\ 
 Ours & \xmark & crop + flip + rot & 51.5 \\
 Ours & \cmark & crop + flip & 58.2 \\
\bottomrule
\end{tabular}
\vspace{0.5em}
 \caption{{\bf Effectiveness of temporal negatives.}
 }
 \label{table:temporal_negatives}
\end{table}

\end{document}